\documentclass[letterpaper,10pt,conference]{ieeeconf}
\IEEEoverridecommandlockouts
\usepackage{amsmath,amssymb,mathtools}
\usepackage{graphicx,subcaption,booktabs,float,placeins}
\usepackage{cite}
\usepackage[hidelinks]{hyperref}
\usepackage[font=small]{caption}
\usepackage{algorithm,algorithmic}
\usepackage{tabularx}
\title{\LARGE\bf
When to Act: Calibrated Confidence for Reliable Human Intention Prediction in Assistive Robotics
}
 
\author{Johannes A.~Gaus$^{1}$, Winfried Ilg$^{1\dagger}$ and Daniel Haeufle$^{1\dagger}$%
\thanks{$^{1}$ Hertie Institute for Clinical Brain Research \& Center for Integrative Neuroscience,
University of Tübingen, Germany. $\dagger$: equal contributing authors}
}

\begin{document}
\bstctlcite{IEEEexample:BSTcontrol}
\maketitle
\thispagestyle{empty}
\pagestyle{empty}

%\begin{abstract}
%Assistive devices must decide not only what a user intends to do but also when a prediction is reliable enough to trigger support. We study calibrated confidence for multimodal intention prediction in activities of daily living using verb level EGTEA~Gaze+ labels that map directly to assistive primitives. A lightweight multimodal GRU achieves about 40\% Top-1 and 70\% Top-5 next action anticipation accuracy under a cross-subject protocol, but its probabilities are strongly overconfident (ECE $\approx0.40$). Post-hoc calibration reduces miscalibration by an order of magnitude (temperature scaling to $0.07$, isotonic regression to $0.04$) without affecting accuracy. When these calibrated confidences drive a simple \textsc{Act}/\textsc{Hold} gate, act only precision increases substantially on the high confidence subset while coverage decreases in a controlled way. This turns the confidence threshold into an interpretable safety parameter and enables reliable assistive triggering. We also outline how calibrated confidence integrates into the iAssistADL control loop \cite{ilg25_icorr}.
%\end{abstract}

\begin{abstract}
Assistive devices must determine both what a user intends to do and how reliable that prediction is before providing support. We introduce a safety–critical triggering framework based on calibrated probabilities for multimodal next–action prediction in Activities of Daily Living. Raw model confidence often fails to reflect true correctness, posing a safety risk. Post–hoc calibration aligns predicted confidence with empirical reliability and reduces miscalibration by about an order of magnitude without affecting accuracy. The calibrated confidence drives a simple \textsc{Act}/\textsc{Hold} rule that acts only when reliability is high and withholds assistance otherwise. This turns the confidence threshold into a quantitative safety parameter for assisted actions and enables verifiable behavior in an assistive control loop.
\end{abstract}

\section{Introduction}
Wrong assistance can be worse than no assistance. In assistive robotics, prediction systems must decide not only \emph{what} a user intends to do but also \emph{when} that prediction is reliable enough to trigger support. This is particularly critical in Activities of Daily Living (ADL), where false assists can confuse users, exacerbate symptoms, or in the worst case lead to injuries. For example, misinterpreting an involuntary tremor as a reach intention could incorrectly activate movement assistance and destabilize the user. Modern deep networks output softmax probabilities that are often interpreted as confidence, yet these values frequently do not match the true likelihood of being correct~\cite{guo2017calibration}. Such miscalibration means that high numerical confidence does not necessarily imply high reliability, creating a safety risk when confidence is used directly for actuation.
The \textsc{Act}/\textsc{Hold} gate must therefore account for human physiological constraints, distinguishing purposeful motion from the inherent unpredictability of pathological signals like ataxia or tremors.

In this work intention refers to short-horizon, verb-level actions such as \emph{reach}, \emph{grasp}, \emph{pour}, or \emph{cut}. The model predicts the next meaningful action that will begin within the next few hundred milliseconds based on a brief multimodal observation window. This level of abstraction lies between long-term task goals and low-level motor predictions, and it aligns directly with the assistive primitives used by the iAssistADL device~\cite{ilg25_icorr}. Verb-level intention is also the correct granularity for triggering assistance because the device implements support profiles at this level rather than at the fine-grained motor or high-level task scale. The device aims to suppress pathological motion in users with disorders such as tremor or ataxia while preserving intended movement. To do this safely, it must detect when the user is initiating a purposeful action and avoid intervening during involuntary or unstable motion. Similar requirements arise in many assistive and shared-control systems that must decide when to intervene under uncertainty.

A confidence-aware intention predictor is therefore needed, one that can abstain when uncertainty is high and act promptly when confidence is trustworthy. We formulate this as a selective prediction problem in which the model outputs both a label and a calibrated confidence, and a downstream \textsc{Act}/\textsc{Hold} gate trades assistance coverage for reliability~\cite{geifman2017selective}.

\begin{figure}[t]
  \centering
  \includegraphics[scale=0.4]{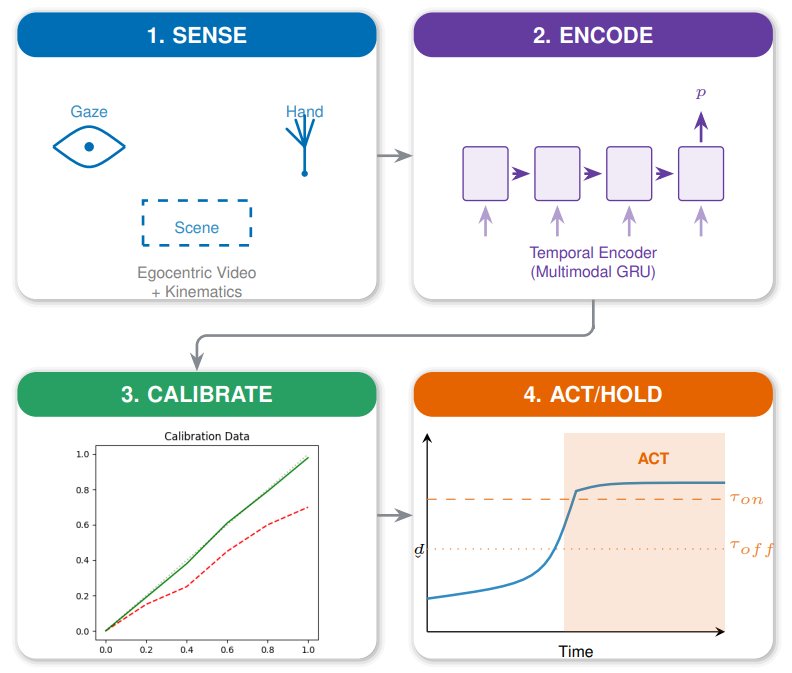}
  \caption{\textbf{Calibrated assistance pipeline.} The system (1) senses multimodal context, (2) encodes intention with a multimodal GRU, (3) calibrates scores so that confidence matches empirical accuracy, and (4) uses a hysteretic \textsc{Act}/\textsc{Hold} gate on $\hat{p}$ to trigger reliable assistance.}
  \label{fig:front_concept_panel}
\end{figure}

Egocentric ADL datasets such as EGTEA~Gaze+ pair first-person video with gaze and hand cues, enabling next-action anticipation before onset~\cite{li2023egteagazeplus}. Prior work has improved accuracy and temporal modeling, yet the reliability of predicted probabilities and their use in explicit assistive decision rules remains underexplored. Because neural networks tend to be overconfident, uncalibrated scores can trigger unintended support. Calibration addresses this by aligning predicted confidence with empirical correctness. Figure~\ref{fig:front_concept_panel} summarizes the resulting calibrated assistance pipeline.
We close this gap in reliable assistive triggering by predicting the user's next action alongside a calibrated uncertainty measure that supports safe decision making. 
%Distinguishing between aleatoric and epistemic uncertainty~\cite{kendall2017what}, we employ post-hoc calibration to produce confidence values that are consistent with observed correctness.

\noindent\textbf{Contributions.} We propose a lightweight pipeline comprising a multimodal GRU, post-hoc calibration, and a hysteresis-based safety gate. First, we show that standard calibration methods like Temperature Scaling, Platt Scaling, and Isotonic Regression reduce Expected Calibration Error (ECE) from $\approx 0.40$ to $0.04$ on EGTEA~Gaze+ without sacrificing accuracy. Second, we use this calibrated signal to drive a safety-oriented \textsc{Act}/\textsc{Hold} controller, transforming the confidence threshold into an interpretable, tunable safety parameter. Finally, we validate that the pipeline is robust to modality loss and meets the real-time constraints of the iAssistADL control loop, bridging the gap between probabilistic perception and verifiable safety in assistive robotics.

\section{Related Work}
\paragraph{Intention and Next Action Prediction in ADL}
Understanding what a user intends to do and when this intention becomes actionable is a central problem in assistive human--robot interaction. Recent surveys review deep learning techniques for action anticipation across different domains and tasks \cite{Zhong2023ASO}. Egocentric ADL datasets couple first person video with gaze and hand cues, enabling anticipation of forthcoming actions before onset. EGTEA~Gaze+ provides synchronized gaze, hand, and action annotations in natural kitchen settings \cite{li2023egteagazeplus}, and large scale corpora such as EPIC-Kitchens and its anticipation challenges extend this paradigm to broader environments \cite{damen2022epickitchens,nasirimajd2023epickitchens100unsuperviseddomainadaptation}. Prior work uses recurrent and transformer based models for temporal context \cite{furnari2019rulstm,girdhar2021avt}.
Complementary approaches in human–object interaction, such as HOIMotion, focus on forecasting human motion using egocentric object information \cite{huHoiMotion2024}. Most studies, however, focus on accuracy or anticipation latency and implicitly treat network scores as calibrated probabilities. A few works explicitly model uncertainty or use it when designing loss functions and evaluation measures for egocentric anticipation \cite{FarhaUncert2019,FunariLeveraging2019}, but they do not study calibrated confidence as a safety signal for assistive triggering.

\paragraph{Uncertainty, Calibration, and Selective Prediction}
Deep networks are often overconfident, which is problematic when scores trigger actions in safety critical systems \cite{guo2017calibration}. Uncertainty work distinguishes aleatoric data ambiguity from epistemic uncertainty due to limited training coverage \cite{kendall2017what,gawlikowski2023uncertainty}, and proposes tools such as Monte Carlo(MC) Dropout, deep ensembles, and lightweight post-hoc calibration schemes \cite{gal2016dropout,lakshminarayanan2017deepensembles,guo2017calibration}.
Selective prediction formalizes abstention on uncertain inputs by trading coverage against accuracy through a confidence threshold \cite{geifman2017selective,geifman2019selectivenet}.
Selective prediction has been explored in wearable and human activity  recognition \cite{roy2021confcalhar}, but applications to assistive robotics remain limited. Shared control requires reliable intent estimates \cite{SelvaggioSurvey2021,JavdaniShared2018}, yet confidence calibration for assistive triggering remains unexplored.
Our work bridges this gap by coupling  calibrated confidence estimates with quantitative safety guarantees for safe  assistive triggering.

%Recent work emphasizes real-time streaming constraints in action anticipation~\cite{furnari2023streaming}  and 
%Recent work leverages multimodal cues such as gaze trajectories and hand-object interactions to improve intention estimates \cite{koppula2016anticipatory}. 
%However, few analyze how reliable uncertainty should govern safe assistive actuation, although similar mechanisms have been successfully deployed in other high-stakes scenarios, such as coupling confidence-calibrated brain signal decoding with runtime safety monitoring to ensure reliable BCI-based control \cite{kim2025gateduncertaintyawareruntimedual}.

\section{Approach}
\label{sec:approach}

\paragraph{Goal}
Given a short multimodal window preceding an action onset, we predict the next action and provide a reliable confidence signal for a safety aware \textsc{Act}/\textsc{Hold} decision. The pipeline is designed for real time operation and consists of an embedded friendly temporal encoder, post-hoc probability calibration, and a confidence gate.

\paragraph{Inputs and Encoder}
\label{subsec:inputs_encoder}
We operate on extracted EGTEA~Gaze+ feature packs that aggregate gaze, hand, and scene cues per timestep into a single vector $x_t\in\mathbb{R}^{D_{\text{feat}}}$. 
Gaze channels provide normalized image-plane coordinates, validity flags, and finite differences. Hand channels provide up to two MediaPipe hand tracks with velocities. Scene channels encode compact HSV histograms. 
All features are $z$ scored using training statistics and invalid timesteps are masked by $u_t\in\{0,1\}$.

Given hidden features $h_t$ and masks $u_t$, we apply masked mean pooling
\[
\bar h=\frac{\sum_{t=1}^T u_t h_t}{\sum_{t=1}^T u_t+\varepsilon}, 
\qquad 
\ell=W\bar h+b,
\]
followed by a softmax over $K$ verb classes. 

To ensure embedded feasibility, we use a compact multimodal GRU encoder. 
At each timestep the concatenated gaze, hand, and scene features pass through a small fully connected layer, then a single-layer GRU with hidden size 256. The pooled hidden state is fed into a linear classifier producing logits in $\mathbb{R}^K$ with $K{=}21$ verbs. This multimodal GRU is the primary model in all experiments, and a small transformer using the same inputs is evaluated as an architectural ablation in Section~\ref{sec:experiments}.

\paragraph{Training Objective}
\label{subsec:training}
We train a single head verb classifier over the collapsed EGTEA label space with $K{=}21$ classes (20 frequent verbs plus an ``other'' class.  
Given logits $\ell\in\mathbb{R}^K$ and class probabilities $p=\mathrm{softmax}(\ell)$, the network is optimized with class weighted cross entropy and optional label smoothing $\epsilon$. Label smoothing replaces each hard target $y_k$ with a mixture of the one hot label and a uniform distribution,
\[
\tilde y_k = (1-\epsilon) y_k + \frac{\epsilon}{K}.
\]
A small smoothing factor proved helpful for stabilizing training on the long tailed verb distribution without harming calibration, as it prevents the model from becoming overconfident on frequent classes.  
We minimize the class weighted cross entropy
\[
\mathcal{L}_{\mathrm{CE}} =
-\!\sum_{k=1}^{K} w_k\, \tilde y_k \log p_k,
\quad
w_k\propto\frac{1}{\mathrm{freq}(k)}.
\]
where $y_k$ is the one hot verb label and $w_k$ upweights rare classes to counter class imbalance.
We use AdamW~\cite{loshchilov2019decoupled} with learning rate $10^{-3}$, weight decay $10^{-2}$, cosine decay, batch size 128, gradient clipping, and early stopping on validation negative log likelihood. To improve robustness to occasional sensor failures we apply modality dropout~\cite{neverova2016moddrop}, which randomly zeros entire modalities such as gaze or hand features on sub batches so that the encoder learns to recover gracefully when individual feature streams are missing.

\paragraph{Calibration}
\label{subsec:uncalib}
The multimodal GRU produces logits $\ell\in\mathbb{R}^K$ and softmax probabilities $p=\mathrm{softmax}(\ell)$. These raw probabilities are often overconfident and do not match the true likelihood of being correct, which poses a safety risk. Post hoc calibration learns a mapping $\mathcal{C}(\cdot)$ on a held out validation set such that the transformed confidence $\hat p=\mathcal{C}(p)$ better reflects empirical accuracy.

To maintain compatibility with real-time embedded deployment, we restrict uncertainty quantification to lightweight post-hoc methods, thus avoiding the prohibitively high runtime overhead of techniques like MC Dropout \cite{gal2016dropout} or deep ensembles \cite{lakshminarayanan2017deepensembles}, which require multiple forward passes. We compare three standard post-hoc calibration approaches. \textit{Temperature Scaling (TS)} rescales the full logit vector by a single scalar $T{>}0$ ($\ell^\prime=\ell/T$) and recomputes probabilities $p^\prime=\mathrm{softmax}(\ell^\prime)$, adjusting the sharpness of the distribution to better match confidence to empirical accuracy \cite{guo2017calibration}. \textit{Platt Scaling} learns a 1D logistic regression on the top logit, outputting an adjusted probability $\hat c = \sigma(a\cdot \ell_{\max} + b)$ fitted on the validation set \cite{platt1999probabilistic}. Finally, \textit{Isotonic Regression} fits a non-parametric, monotone curve that maps the raw confidence to its empirical accuracy without assuming a parametric form \cite{ZadroznyIsot2002}. Since selective prediction only depends on the maximum class probability, we calibrate only the top class confidence, which avoids unnecessary distortion of the full distribution while preserving the signal used for actuation.
In all cases, the predicted class $\hat y$ remains unchanged; only the numerical confidence value changes, which is then used to drive the \textsc{Act}/\textsc{Hold} gate in Section~\ref{sec:approach} (\nameref{subsec:gate}).

\paragraph{Safety-Oriented Act/Hold Gate}
\label{subsec:gate}
At deployment we use the calibrated probabilities $\hat p$ to drive a binary \textsc{Act}/\textsc{Hold} decision. Let $\hat c = \max_k \hat p_k$ be the calibrated top class confidence.
The basic rule is
\begin{equation}
\boxed{
\textsc{Act}
\;\Leftrightarrow\;
\hat c \ge \tau
}
\label{eq:gate}
\end{equation}
with threshold $\tau\!\in[0,1]$.
Increasing $\tau$ reduces the fraction of windows on which the system acts (coverage) and increases the fraction of correct predictions while in \textsc{Act} (act only precision, AOP), so $\tau$ is the main safety knob.

\paragraph{Stability} 
To prevent rapid switching or "chattering" in confidence borderline regions, we implement a small hysteresis band and a refractory period (\autoref{fig:front_concept_panel}). Assistance is only triggered (\textsc{Act}) when the calibrated confidence $\hat c$ crosses an upper threshold $\tau_{\mathrm{on}}$, and it is held or turned off (\textsc{Hold}) only when $\hat c$ drops below a lower threshold $\tau_{\mathrm{off}}$ ($\tau_{\mathrm{on}} > \tau_{\mathrm{off}}$). After any switch, the gate ignores further changes for a short refractory time $R$\,ms. In the iAssistADL control loop, this mechanism is essential for stability and prevents short unintended bursts of assistance.

\paragraph{Calibration and a safety bound}
If the calibrated confidences are accurate on the decision region $\{\hat c\ge\tau\}$, in the sense that the calibration error there is bounded by $\varepsilon$ (binned confidence differs from empirical accuracy by at most $\varepsilon$), then the act only precision satisfies
\begin{align*}
\mathrm{AOP}(\tau)
&=\mathbb{E}\!\left[\mathbf{1}\{\hat y{=}y^\star\}\mid \hat c\ge\tau\right] \\
&\ge \mathbb{E}\!\left[\hat c\mid \hat c\ge\tau\right]-\varepsilon \\
&\ge \tau-\varepsilon.
\end{align*}
The inequality states that when the system acts only on windows with confidence at least $\tau$, the probability of being correct on those windows cannot fall much below $\tau$ itself, up to the calibration error. Better calibration (smaller $\varepsilon$) therefore sharpens this guarantee and turns $\tau$ into a meaningful lower bound on the reliability of all assisted actions, providing a simple quantitative safety control for triggering assistance.

\paragraph{Dataset, Task, Splits, and Preprocessing}
\label{sec:data_impl}
We evaluate on EGTEA Gaze+ \cite{li2023egteagazeplus}, a 32 subject egocentric ADL dataset with synchronized video, gaze, and action annotations in kitchen settings. Following the standard anticipation protocol, the task is next action anticipation: given a short multimodal window before the onset of an annotated action, the model predicts the next action label $y$ and a confidence signal that can drive a safety gate.

The raw annotations contain several hundred verb noun strings. We focus on verb level intention, which is more directly aligned with assistive control primitives such as reach, pour, or cut. Verbs are normalized (lowercasing, typo correction, inflection merging) and collapsed to 20 frequent verbs plus a residual other class, giving $K{=}21$ classes. The other class aggregates many rare verbs and is therefore larger than most individual classes, reflecting the long tailed distribution of EGTEA actions. This verb vocabulary is used consistently for training, calibration, and evaluation.

We use subject disjoint 24/4/4 train/val/test splits across three folds. Sequences are cut into 2 s windows ending 0.5 s before action onset. We selected this window duration because it captures preparatory motion without leaking onset frames, consistent with prior ADL anticipation protocols \cite{furnari2019rulstm}. Each window contains pre extracted gaze, hand, and scene features resampled to 25 Hz and normalized per fold; windows with insufficient valid samples are discarded during training. Verb labels follow the same 21 class space.

\section{Experiments \& Results}
\label{sec:experiments}
We evaluate calibrated multimodal intention predictors on EGTEA~Gaze+ under the cross-subject protocol of Section~\ref{sec:approach} (\nameref{sec:data_impl}).
Unless stated otherwise, inputs are gaze+hand+scene, the default encoder is the multimodal GRU from Section~\ref{sec:approach} (\nameref{subsec:inputs_encoder}), and results are averaged over three subject-disjoint folds.

\subsection{Baseline Accuracy in Verb Level EGTEA Gaze+ Anticipation}
\label{subsec:baseline_acc}
Before analyzing calibration and selective prediction, we first establish baseline accuracy for our multimodal models on the verb level EGTEA~Gaze+ anticipation task. This provides the reference performance against which all subsequent calibration and safety results are interpreted.

On EGTEA~Gaze+, Furnari et al.\ report that RU-LSTM, which uses high capacity RGB, optical flow and object features with modality attention, achieves Top-1 action recognition of 33.06\% and 19.49\% on the two official test sets when the entire action segment is observed in the full 106-class verb--noun space \cite{furnari2019rulstm}. Direct comparison to our work is difficult due to the different action granularities (106 vs.\ 21 classes), observation windows (full action vs.\ 2\,s anticipation), and feature types (end-to-end RGB vs.\ multimodal). Anticipation with only a partial observation is substantially harder \cite{furnari2019rulstm}, and accuracy drops accordingly. 

In our anticipation setup we observe only a 2.0\,s multimodal window ending 0.5\,s before action onset and we use compact precomputed gaze, hand, and scene features instead of end-to-end RGB encoders. This deliberately trades some absolute recognition performance for embedded feasibility and more predictable calibration, avoiding the compute and memory footprint of large vision backbones on the assistive device.  
 
Under this configuration the multimodal GRU baseline achieves average Test Top-1 accuracy of $0.402\pm0.004$ and Top-5 accuracy of $0.699\pm0.006$ across three subject-disjoint folds. A transformer encoder on the same inputs attains comparable Top-1 accuracy ($0.394\pm0.009$) and slightly higher Top-5 accuracy ($0.709\pm0.028$), confirming that the task is not bottlenecked by temporal capacity but by the intrinsic ambiguity of short-horizon egocentric anticipation.  

These baselines provide a realistic basis for our main goal, which is to study how calibrated confidence and selective prediction can make such anticipatory models usable in a safety-aware assistive setting.

\subsection{E1: Calibration Methods}
\label{subsec:e1_calibration}

\begin{figure}[h]
  \centering
  \includegraphics[scale=0.275]{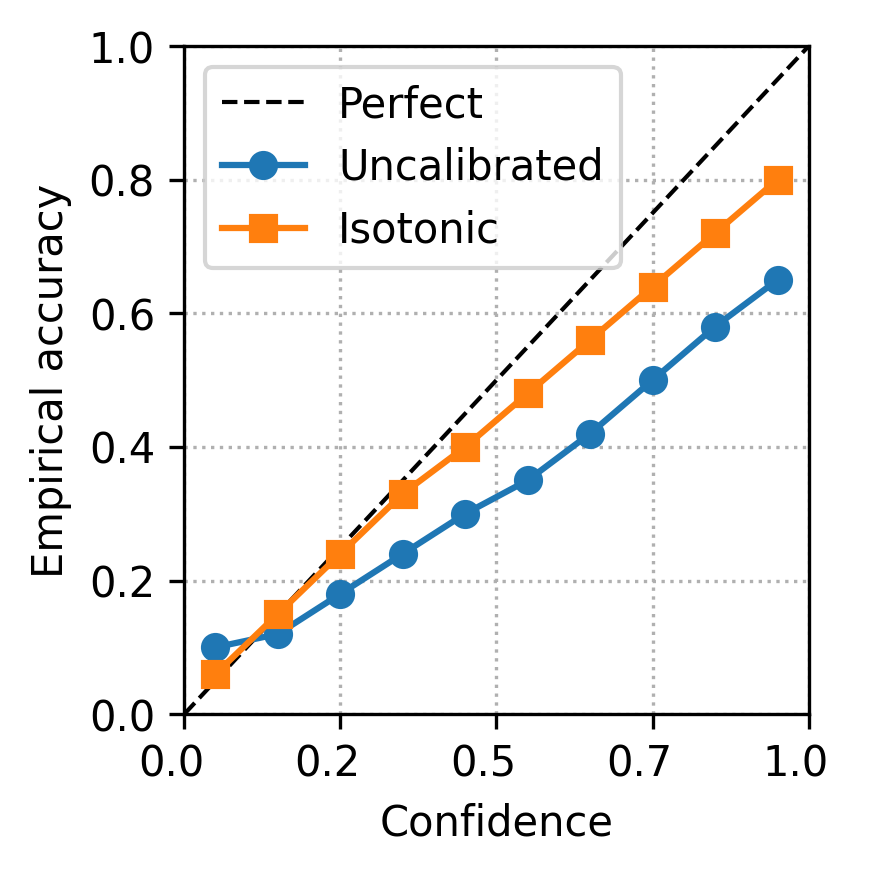}
  \caption{Reliability diagram showing how calibration aligns predicted confidence with empirical accuracy. Isotonic regression follows the identity line more closely than the uncalibrated model.}
  \label{fig:reliability}
\end{figure}

We first quantify how post-hoc calibration affects probability reliability of the multimodal GRU on the verb level task. Table~\ref{tab:e1_calibration} summarizes Top-1 accuracy and calibration metrics for several post-hoc schemes fitted on the validation logits. The uncalibrated model is strongly overconfident (ECE $\approx 0.40$) despite reasonable Top-1 accuracy. Temperature scaling, which optimizes a single global temperature $T$ on validation NLL, substantially reduces miscalibration (ECE $=0.071$) while leaving accuracy unchanged. Isotonic regression on the top-class confidence further lowers ECE to $0.039$, a tenfold reduction compared to the uncalibrated scores, again without affecting accuracy.

\begin{figure}[t]
  \centering
  \includegraphics[scale=0.85]{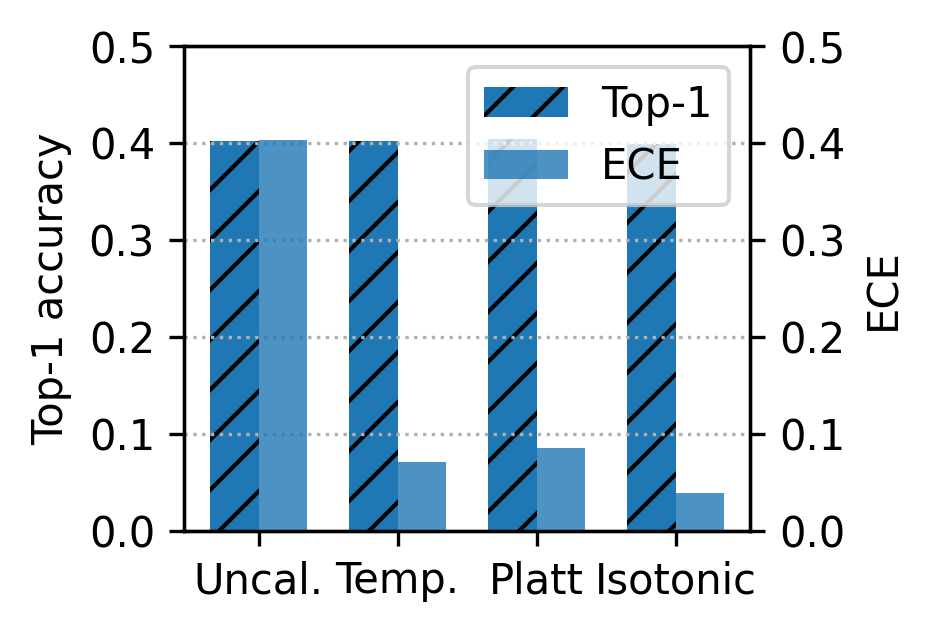}
  \caption{Calibration performance across methods. Temperature scaling and isotonic regression reduce ECE substantially while leaving Top-1 accuracy unchanged.}
  \label{fig:calibration_bar}
\end{figure}

Platt scaling yields similar accuracy and improves both NLL and Brier score compared to the uncalibrated model. Temperature scaling reduces both metrics (NLL from 2.73 to 2.22, Brier from 0.964 to 0.785), and Platt scaling pushes them slightly lower still, although its ECE remains higher than TS. Isotonic regression provides the strongest top class calibration (ECE $0.039$), but because it calibrates only the top class confidence, NLL and Brier scores are not directly comparable. For selective prediction, where the controller depends solely on the maximum class probability, top class isotonic regression provides the most relevant reliability improvement while avoiding distortion of the full distribution. Figure~\ref{fig:calibration_bar} visualizes these effects on ECE and NLL across methods.

\begin{table}[h]
\begin{center}
\caption{Calibration performance of the multimodal GRU on verb level EGTEA~Gaze+. Values are averaged over three folds; ECE is reported in absolute units.}
\label{tab:e1_calibration}
\begin{tabular}{lcccc}
\toprule
Method & Top-1 $\uparrow$ & ECE $\downarrow$ & NLL $\downarrow$ & Brier $\downarrow$ \\
\midrule
Uncalibrated                & 0.402 & 0.403          & 2.73 & 0.964 \\
Temperature scaling         & 0.402 & 0.071          & 2.22 & 0.785 \\
Platt scaling               & 0.404 & 0.085          & 2.06 & 0.753 \\
Isotonic (top class)$^\dagger$ & 0.399 & \textbf{0.039} & n/a  & n/a  \\
\bottomrule
\end{tabular}
\end{center}
\vspace{1pt}
\footnotesize
$^\dagger$Isotonic regression is applied only to the top-class confidence, so full-distribution NLL and Brier scores are not directly comparable and are therefore omitted.
\end{table}

The reliability diagram in Fig.~\ref{fig:reliability} further illustrates how isotonic regression aligns predicted confidence with empirical correctness.

\subsection{E2: Modality Ablations and Robustness Snapshot}
\label{subsec:e2_uncert}

To assess robustness under degraded sensing, we evaluate the multimodal GRU when individual input streams are removed at test time. Removing either hand or scene features causes accuracy to collapse to near chance level, %(Fig.~\ref{fig:modality_ablation}) 
confirming that these two modalities carry most of the discriminative signal for verb level anticipation.

%Dropping hand or scene features collapses performance to chance level (Fig.~\ref{fig:modality_ablation}), confirming their central role in verb level anticipation.

We also compute a lightweight diagonal Laplace approximation \cite{daxberger2021laplace} to obtain a coarse estimate of parameter sensitivity. Curvature concentrates almost entirely in the final feature projector and classification head, while most recurrent parameters lie in a flat region. This suggests that simple Bayesian post-processing of the head could capture most epistemic uncertainty without additional runtime cost.
%\begin{figure}[h]
%  \centering
%  \includegraphics[scale=0.85]{img/fig_modality_ablation.png}
%    \caption{Modality ablation. Removing hand or scene features collapses accuracy to near-chance, indicating their dominant role in verb level anticipation.}
%  \label{fig:modality_ablation}
%\end{figure}
Overall, the model degrades predictably under modality loss and shows a clear separation between informative channels (hand, scene) and less informative channels (flow, semantic). This predictable behavior is important for real-world assistive sensing pipelines, where intermittent sensing failures are common and cross-modal redundancy helps maintain stable performance and safe operation.

\subsection{E3: Safety Decision Analysis}
\label{subsec:e3_safety}

We study selective prediction by sweeping a confidence threshold $\tau$ on the calibrated top class probability and measuring act only precision (AOP) versus coverage. At $\tau=0$ the multimodal GRU acts on every window and recovers the baseline Top-1 accuracy of about 40\%. Increasing $\tau$ forces the system to act only when its confidence is high. This reduces coverage but increases AOP, because the model tends to be correct on the subset of windows where it is most certain.

Unlike safety-critical autonomy where missing a detection is catastrophic, in this shared-control assistive setting we strictly prioritize act-only precision over coverage; a false positive injects active forces that could physically destabilize a user, whereas a missed assist simply defaults the device to a safe, transparent ``follow'' mode.

This is the core idea behind using confidence as a safety filter. A low threshold $\tau$ corresponds to a high availability mode and a higher threshold corresponds to a conservative mode where almost every triggered assist is correct. Calibration improves this trade off: at any fixed coverage level the calibrated model achieves higher AOP, showing that well behaved probabilities translate directly into safer operating regimes for the \textsc{Act}/\textsc{Hold} gate.

This empirical trend matches the theoretical bound in Section~\ref{sec:approach} (\nameref{subsec:gate}). When calibration error is small on the decision region, a threshold $\tau$ guarantees that assisted actions have precision at least $\tau$ minus a small margin. Better calibration therefore sharpens this guarantee and makes the threshold an interpretable safety parameter rather than a heuristic.

\subsection{E4: Runtime and Embedded Feasibility}
\label{subsec:e4_runtime}

All experiments use pre extracted multimodal features, so test time inference only runs the multimodal GRU and a small linear head. A forward pass for a typical window of $T{\approx}50$ timesteps requires about $2$–$3$ ms on CPU, which fits comfortably within a 40 ms sensing and control cycle typical for real time assistive devices, including iAssistADL. The feature extraction chain comprising gaze parsing, MediaPipe Hands, and HSV histograms can run in real time at 25 Hz on modest embedded hardware, making the full pipeline suitable for on device deployment in a range of assistive platforms.

During operation the intention predictor processes a sliding window at 25 Hz and outputs the calibrated maximum confidence $\hat c$. This confidence directly drives the binary \textsc{Act}/\textsc{Hold} signal used by the motion controller. In \textsc{Hold} the system follows the user without assistance, while in \textsc{Act} it blends the corresponding assistive profile into the ongoing movement. To prevent rapid switching, the gate employs a small hysteresis band with thresholds $\tau_{\mathrm{on}}$ and $\tau_{\mathrm{off}}$ and applies a short refractory period after each transition. This stabilizes the closed loop and suppresses brief unintended assists. Because calibration aligns $\hat c$ with empirical correctness, the threshold $\tau$ becomes an interpretable safety parameter that trades availability against reliability and can be tuned in any assistive controller that consumes a confidence signal.

\subsection{E5: Online Stream Simulation with a smoothed confidence gate}
\label{subsec:closed_loop}

\begin{figure}[h]
  \centering
  \includegraphics[scale=0.85]{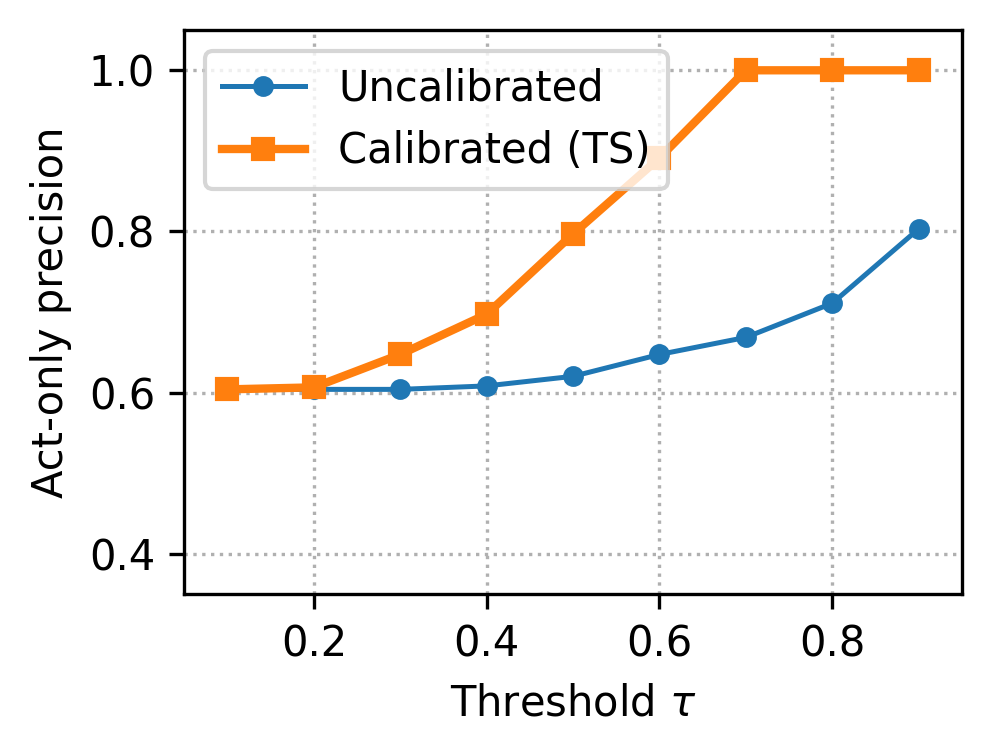}
  \caption{Closed-loop act-only precision as a function of the confidence threshold $\tau$. Calibration turns the threshold into a meaningful safety parameter: the calibrated gate achieves high precision already at mid-range $\tau$, while the uncalibrated scores remain flat until extreme thresholds.}
  \label{fig:precision_vs_tau}
\end{figure}

To approximate the behavior of the intention module inside a running assistive device, we perform an offline closed-loop simulation by replaying per-window predictions along the recorded EGTEA~Gaze+ sequences and driving a discrete \textsc{Act}/\textsc{Hold} state machine. This simulation uses exactly the same calibrated confidence signal, hysteresis band, and refractory \textsc{Act}/\textsc{Hold} logic that we intend to run on the iAssistADL controller, so the only missing component in our experiments is the physical actuator. For each fold, we load validation logits and targets, apply temperature scaling or leave the scores uncalibrated, and run this closed-loop simulation with:
(i) exponential smoothing of class probabilities with factor $\alpha{=}0.2$,  
(ii) a top-$k$ filter with $k{=}3$ that only considers windows where the assist verb is among the three most likely classes, and  
(iii) a confidence threshold $\tau$ that switches the gate between \textsc{Act}$\,$/$\,$\textsc{Hold}.

For each $\tau$ we measure the fraction of time spent in \textsc{Act} (coverage) and the fraction of correct actions while in \textsc{Act} (act-only precision). At $\tau{=}0$, the system acts on all windows and recovers the baseline Top-1 accuracy of about $0.40$. As $\tau$ increases, coverage drops while act-only precision rises.

Figure~\ref{fig:precision_vs_tau} shows that calibration transforms the confidence threshold into a usable safety dial: at $\tau=0.5$ act-only precision increases from $0.62$ (uncalibrated) to $0.80$ (calibrated), a 47\% relative improvement at less than half the coverage.

With uncalibrated scores the low-threshold regime is almost flat and several different cutoffs produce the same behavior, so the threshold is not an interpretable safety parameter.
Typical operating points include: a balanced mode with coverage around $0.5$ and precision around $0.65$, and a conservative mode with coverage around $0.25$ and precision close to $0.80$.

These simulations show that post-hoc calibration does not increase raw accuracy but is crucial for making confidence thresholds behave as meaningful, fold-stable safety parameters in a closed-loop setting.

\section{Conclusion}
\label{sec:conclusion}

The central claim of this paper is that probability reliability is more critical than raw accuracy for safe assistive triggering. On EGTEA~Gaze+, our lightweight multimodal GRU achieved $\approx 40\%$ Top 1 accuracy, similar to a comparison Transformer, indicating that short horizon ambiguity in the data rather than model capacity is the primary bottleneck. Modality ablations further show that hand and scene features dominate prediction, while the gaze channel provides a moderate boost and degrades gracefully when removed.

Raw model confidence is, however, severely overconfident (ECE $\approx 0.40$). Post hoc calibration with Isotonic Regression reduces this miscalibration to $0.04$ without affecting accuracy, transforming the confidence score into a trustworthy control signal. Selective prediction driven by this calibrated confidence consistently improves the accuracy–coverage curve, validating the confidence driven \textsc{Act}/\textsc{Hold} gate.

From a safety perspective, the calibrated gate directly addresses the balance between false assists and missed assists. Aligning numerical scores with empirical correctness makes the threshold $\tau$ an explicit, interpretable safety knob: increasing $\tau$ yields conservative high precision behavior, while decreasing it increases availability, enabling patient specific tuning. This alignment also supports established risk management practices, allowing thresholds to function as risk control measures and strengthening the link between probabilistic perception and device safety requirements.

\noindent\textbf{Limitations and Outlook:} Our evaluation is based on offline replay of recorded sequences, necessitating prospective validation with real users in a closed loop setting, such as the iAssistADL device, to verify safety in deployment. The smooth kinematic profiles in EGTEA~Gaze+ also differ from pathological motion in our target demographic (e.g., ataxia or tremor), motivating tests of calibration stability under such domain shifts and potentially patient specific online recalibration. While we focused on verb level intentions to simplify the action space and improve calibration stability, future work should extend the approach to fine grained verb noun prediction and incorporate online recalibration to maintain reliability across new environments and populations. Despite these limitations, the results confirm that calibrated confidence provides a practical, safe boundary for assistive triggering that meets real time constraints with negligible overhead.

\section*{Acknowledgments}
This work was financed by the Baden-Württemberg Stiftung in the scope of the AUTONOMOUS ROBOTICS project \textit{iAssistADL} granted to DH and WI.

\bibliographystyle{IEEEtran}
\bibliography{references}

\end{document}